%% file: main.tex
\documentclass[runningheads]{llncs}

\usepackage{eccv}

\input{preamble.tex}
\usepackage{eccvabbrv}

\usepackage{graphicx}
\usepackage{booktabs}
\usepackage{color,soul}

\usepackage[accsupp]{axessibility}  

\usepackage{hyperref}

\usepackage{orcidlink}
\usepackage{multirow}

\begin{document}

\title{RoWeeder: Unsupervised Weed Mapping through Crop-Row Detection}

\titlerunning{RoWeeder}

\author{Pasquale De Marinis\orcidlink{0000-0001-8935-9156} \and
Gennaro Vessio\orcidlink{0000-0002-0883-2691} \and
Giovanna Castellano\orcidlink{0000-0002-6489-8628}}

\authorrunning{P.~De Marinis et al.}

\institute{University of Bari Aldo Moro, Bari, Italy\\
\email{\{pasquale.demarinis,gennaro.vessio,giovanna.castellano\}@uniba.it}
}

\maketitle

\begin{abstract}
Precision agriculture relies heavily on effective weed management to ensure robust crop yields. This study presents RoWeeder, an innovative framework for unsupervised weed mapping that combines crop-row detection with a noise-resilient deep learning model. By leveraging crop-row information to create a pseudo-ground truth, our method trains a lightweight deep learning model capable of distinguishing between crops and weeds, even in the presence of noisy data. Evaluated on the WeedMap dataset, RoWeeder achieves an F1 score of 75.3, outperforming several baselines. Comprehensive ablation studies further validated the model's performance. By integrating RoWeeder with drone technology, farmers can conduct real-time aerial surveys, enabling precise weed management across large fields. The code is available at: \url{https://github.com/pasqualedem/RoWeeder}.
\footnote{Presented at the Computer Vision for Plant Phenotyping and Agriculture (CVPPA) workshop at ECCV 2024.}
\end{abstract}

\input{sec/paper}

\bibliographystyle{splncs04}
\bibliography{main}
\end{document}

%% file: preamble.tex
\usepackage[dvipsnames]{xcolor}

\usepackage{algorithm}
\usepackage{algpseudocode}
\usepackage{svg}
\usepackage{tensor}

%% file: sec/paper.tex
\section{Introduction}
\label{sec:introduction}

Agriculture is essential for human sustenance, and advancements in farming machinery and techniques have improved crop yield. \textit{Weed management} is crucial for removing unwanted plants that compete with crops. Effective weed management enhances crop productivity and promotes sustainable agriculture.

Drones, or unmanned aerial vehicles (UAVs), have proven invaluable assets in precision agriculture, offering both versatility and cost-effectiveness~\cite{vougioukas2019}. These devices can capture high-resolution imagery and data from farmlands, enabling farmers to monitor crop development, detect diseases and pests, and fine-tune irrigation strategies. By providing accurate and timely information, drones help reduce expenses, boost crop yields, and minimize the use of inputs such as water, fertilizers, and pesticides. Traditional methods like manual field inspections or satellite-based remote sensing cannot match the level of detail drones provide. UAVs can conduct real-time aerial surveys of crops, allowing farmers to make prompt, informed decisions. Moreover, drones can efficiently cover vast areas in a matter of hours, a task that would typically require days or weeks using conventional approaches. This time-saving capability enables farmers to conserve resources and make more timely decisions.

Recent research has suggested that deep learning models can be used for semantic segmentation and weed identification in drone-captured or aerial imagery~\cite{dos2017weed,sa2017weednet,castellano2023weed}. However, despite these significant advancements, automatically detecting weeds remains a complex challenge. Deep learning techniques are not yet widely adopted in agriculture, mainly due to the extensive manual annotation required for the large amount of data needed in the learning phase. This issue is particularly pronounced in agricultural datasets, where labeling plants in field images is time-consuming. Furthermore, these models often demand significant computational resources, posing challenges for drone implementation with constrained processing capabilities and limited power supply. The need for real-time performance, coupled with these limitations, underscores the importance of developing \textit{lightweight} solutions for weed mapping applications.

In agricultural landscapes, crops are systematically planted in linear crop rows. Automatically detecting these crop rows can benefit various applications \cite{stefanovicBlueberryRowDetection2023, waqarEndtoEndDeepLearning}. Among them, it can be used to distinguish crops from weeds, as unwanted vegetation typically proliferates in the spaces between these rows. This spatial pattern has led to a significant body of research focused on developing and refining \textit{crop-row} detection techniques~\cite{guerreroAutomaticExpertSystem2013,vidovicCropRowDetection2016,jiCroprowDetectionAlgorithm2011,bahCRowNetDeepNetwork2020,bahHierarchicalGraphRepresentation2023}. This mechanism allows to detect inter-rows weeds. However, it may fail to detect intra-row weeds, as they are classified as crops. The generated detection can be used to create a pseudo-ground truth, which can be used to train a deep learning model. This approach has been explored in previous works~\cite{bahDeepLearningUnsupervised2018,bahDeepFeaturesOneclass2019a}, but these methods primarily functioned as image classifiers rather than end-to-end segmentation tools.

This paper proposes a novel lightweight, fully automatic method for weed mapping that combines crop-row detection with a \textit{noise-resilient} deep learning model. Our approach, RoWeeder (Fig.~\ref{fig:framework}), leverages the crop-row information to create a pseudo-ground truth, which is then used to train a deep learning model. The model is based on the SegFormer encoder architecture \cite{xie2021segformer}, coupled with an ad-hoc decoder that fuses the features extracted by the encoder. Evaluation is conducted on the WeedMap dataset~\cite{sa2018weedmap}, which contains multispectral images of sugar beet fields captured by drones.

The rest of this paper is organized as follows. Section~\ref{sec:related_work} reviews related work. Section~\ref{sec:roweeder} details our proposed approach. Section~\ref{sec:experiments} evaluates our model's performance. Section~\ref{sec:conclusion} concludes the paper with a summary of our findings and directions for future research.

\section{Related Work}
\label{sec:related_work}

Advances in computer vision and remote sensing have revolutionized precision agriculture, addressing tasks such as disease and pest identification, abiotic stress assessment, growth monitoring, crop yield prediction, and weed mapping.

Weed mapping, a semantic segmentation task, assigns each pixel in an image to one of two classes, either \textit{weed} or \textit{crop}. Deep learning algorithms have demonstrated superior performance over traditional techniques in this area. Early work by Dos Santos et al.~\cite{dos2017weed} highlighted the effectiveness of Convolutional Neural Networks (CNNs), such as AlexNet, over SVMs and Random Forests. Lottes et al.~\cite{lottes2018joint} further advanced the field by using a CNN with dual decoders for stem detection and plant segmentation, showing promising results on the BoniRob and UAV datasets.

Integrating multispectral images, which capture detailed information on plant health and species, enhances the accuracy of deep learning models compared to RGB-only models. For example, U-Net has successfully separated weeds from crops and soil~\cite{chicchon2019semantic}. WeedNet, based on SegNet and trained on the WeedMap dataset, is another example of successful application in this domain~\cite{sa2017weednet}. The WeedMap dataset, containing multispectral images from sugar beet fields in Germany and Switzerland, has become a benchmark for weed mapping studies~\cite{sa2018weedmap}. Recently, we have explored the benefits of RGB pre-training and fine-tuning on multispectral images, as well as the application of knowledge distillation to improve the performance of lightweight models~\cite{castellano2023weed,castellano2023applying}. 

Partially supervised methods, such as semi supervised and unsupervised learning, have also been explored for weed detection. Studies employing semi supervised approaches include~\cite{perez-ortizSemisupervisedSystemWeed2015,nongSemisupervisedLearningWeed2022a,khanNovelSemisupervisedFramework2021,shorewalaWeedDensityDistribution2021a}, while clustering methods for unsupervised weed detection are discussed in~\cite{gasparovicAutomaticMethodWeed2020,agarwalWeedIdentificationUsing2021,peerbhayImprovingUnsupervisedMapping2021}. Crop-row detection methods, such as those using the Hough Transform, have been employed to distinguish between crop plants and weeds~\cite{penaWeedMappingEarlySeason2013, dossantosferreiraUnsupervisedDeepLearning2019a, perez-ortizSelectingPatternsFeatures2016a}. Bah et al.~\cite{bahDeepLearningUnsupervised2018, bahDeepFeaturesOneclass2019a} introduced deep learning techniques incorporating crop-row information, though these methods primarily functioned as image classifiers rather than end-to-end segmentation tools.

Building on these unsupervised methods, our work proposes a novel approach that combines crop-row detection with a noise-resilient end-to-end deep learning model. This lightweight model can detect weeds in real-time during inference, integrating crop-row information learned during training. To our knowledge, this is the first work that merges these two approaches.

\section{RoWeeder}
\label{sec:roweeder}

\begin{figure}[t]
    \centering
    \includegraphics[width=\linewidth]{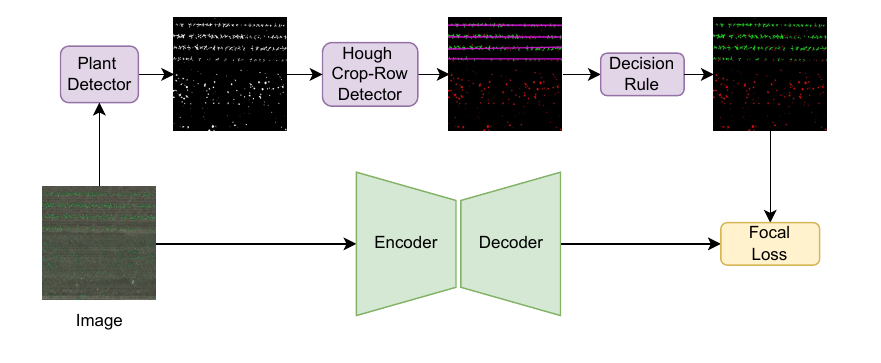}
    \caption{Overview of the proposed framework, illustrating the training process of the semantic segmentation model. In the top branch, the image is fed into a plant detection algorithm, and the mask produced is used to detect crop rows. Every plant on a crop row is classified as a crop; otherwise, it is classified as a weed. This pseudo-ground truth is used to train a deep learning model.}
    \label{fig:framework}
\end{figure}

This research aims to develop a framework that can transfer knowledge derived from crop-row detection to a deep learning model for weed mapping. Our approach, RoWeeder, uses crop-row information to create a pseudo-ground truth, which is then used to train the deep learning model. The proposed framework consists of three main components: a plant detection module, a crop-row detection module, and a deep learning model for final segmentation. Figure~\ref{fig:framework} illustrates the workflow of the RoWeeder framework.

The plant detection module uses classical thresholding to detect plants in the image. When Near Infrared (NIR) images are available, the Normalized Difference Vegetation Index (NDVI) is calculated to improve plant detection.

The crop-row detection module employs the Hough Transform \cite{dudaUseHoughTransformation1972} to detect crop rows in the image. The module takes the segmentation mask of the plants as input and outputs the crop-row mask. The crop-row mask is then used to create the pseudo-ground truth for the deep learning model. Since all the images  
are rotated by the same angle, they are first rotated to align the crop rows with the horizontal axis before crop-row detection. This consistent rotation ensures accurate detection by the Hough Transform. Moreover, the data may contain images filled with weeds (found at the edges of the fields), leading to the detection of false positive lines by the Hough Transform. In these images, the detected lines' angles ($\theta$) are evenly distributed, while in images with crop rows, the $\theta$ angles are concentrated around a specific value. To tackle this issue, we analyze the distribution of $\theta$ angles to filter out false positive lines. We use the Kolmogorov-Smirnov test to check the uniformity of the $\theta$ angle distribution, and if the distribution is found to be uniform, we discard all the detected lines.

\begin{figure}[t]
    \centering
    \begin{subfigure}{0.23\linewidth}
      \includegraphics[width=\linewidth]{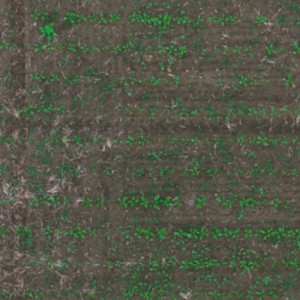}
    \end{subfigure}
    \begin{subfigure}{0.23\linewidth}
      \includegraphics[width=\linewidth]{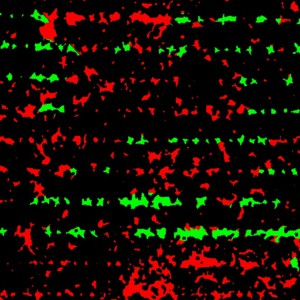}
    \end{subfigure}
    \begin{subfigure}{0.23\linewidth}
      \includegraphics[width=\linewidth]{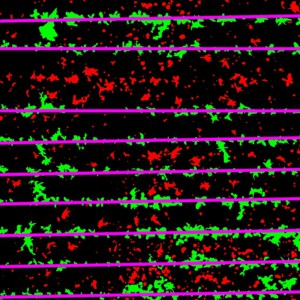}
    \end{subfigure}

    \begin{subfigure}{0.23\linewidth}
        \includegraphics[width=\linewidth]{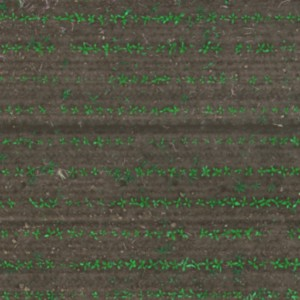}
      \end{subfigure}
      \begin{subfigure}{0.23\linewidth}
        \includegraphics[width=\linewidth]{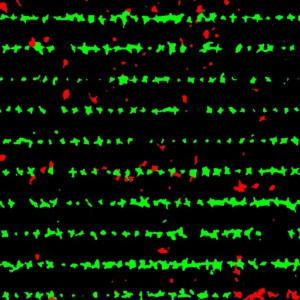}
      \end{subfigure}
      \begin{subfigure}{0.23\linewidth}
        \includegraphics[width=\linewidth]{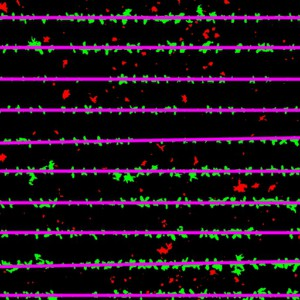}
    \end{subfigure}

    \caption{Visual representation of the crop lines detected by the Hough Transform. From left to right: the original input image, the ground truth with crops in green and weeds in red, and the generated pseudo-ground truth with detected crop rows in purple.}
    \label{fig:hough}
\end{figure}

Crop and weed classification is performed using a simple rule-based method: if at least one pixel of a plant overlaps with a crop row, it is classified as a crop; otherwise, it is classified as a weed. Each plant instance can be detected using either the Simple Linear Iterative Clustering (SLIC) algorithm~\cite{achantaSLICSuperpixels2010} or by calculating each connected component (CC) of the image. This method effectively detects inter-row weeds but may fail to detect intra-row weeds, as they are classified as crops. The hypothesis is that the deep learning model will learn to distinguish between crops and weeds despite the pseudo-ground truth noise. Examples of segmentations produced by the Hough Crop-Row Detector with the decision rule are shown in Fig.~\ref{fig:hough}.

RoWeeder is built on the SegFormer architecture~\cite{xie2021segformer}, a Transformer-based model that has shown state-of-the-art performance in semantic segmentation tasks. SegFormer comes in six configurations, ranging from small to large models. We chose the smallest configuration, SegFormer-B0, which has only 3.7M parameters, to ensure that the model is light enough to be deployed on edge devices like NVIDIA Jetson TX2 and Jetson Nano \cite{leeLightweightMonocularDepth2023}. The model is trained on the pseudo-ground truth created by the crop-row detection module.

SegFormer employs an all-MLP lightweight decoder. We designed different segmentation decoders to identify the most suitable for our task. The first is a pyramid-based decoder. Starting from the deepest feature map, at each stage, this map is upsampled to the next one, projected to the same dimension as the feature map with a point-wise convolution, and then fused. Fusion can be done by concatenation or addition, while upsampling can be done by bilinear interpolation or transposed convolution. Finally, the fused feature map is processed by a $3 \times 3$ convolution to handle spatial information, followed by a GELU activation function~\cite{hendrycksGaussianErrorLinear2023}. Given $F_{deep}$ and $F_{shallow}$, the deepest and shallowest feature maps, a pyramid-based decoder block with addition as fusion can be defined as:
\begin{equation}
F_{out} = GELU\left(\text{Conv}_{3\times3}\left(\text{Conv}_{1\times1}\left(\text{Upsample}\left(F_{deep}\right)\right) + F_{shallow}\right)\right)
\end{equation}
where $\text{Upsample}$ can be bilinear interpolation or transposed convolution. A pyramid-based decoder block with concatenation as fusion can be defined as:
\begin{equation}
F_{out} = GELU\left(\text{Conv}_{3\times3}\left(\text{Conv}_{1\times1}\left(\text{Upsample}\left(F_{deep}\right) || F_{shallow}\right)\right)\right)
\end{equation}
where $||$ denotes concatenation. 

The second decoder is a ``flat'' decoder that projects all the feature maps in one step. The flat decoder with sum as fusion can be defined as:
\begin{equation}
F_{out} = GELU\left(\text{Conv}_{3\times3}\left(\sum_{n=1}^{N}(\text{Conv}_{1\times1}(\text{Upsample}(a_n)))\right)\right)
\end{equation}
where $a_n$ is the $n$-th feature map, $N$ is the number of blocks, $\text{Upsample}$ can be either bilinear interpolation or transposed convolution, and $\text{Fusion}$ can be either concatenation or addition.
The flat decoder with concatenation as fusion can be defined as:
\begin{equation}
F_{out} = GELU\left(\text{Conv}_{3\times3}(\text{Conv}_{1\times1}(\text{Upsample}(a_1) || \ldots || \text{Upsample}(a_N)))\right)
\end{equation}
where $||$ denotes concatenation.

\begin{figure}[t]
    \centering
    \includegraphics[width=\linewidth]{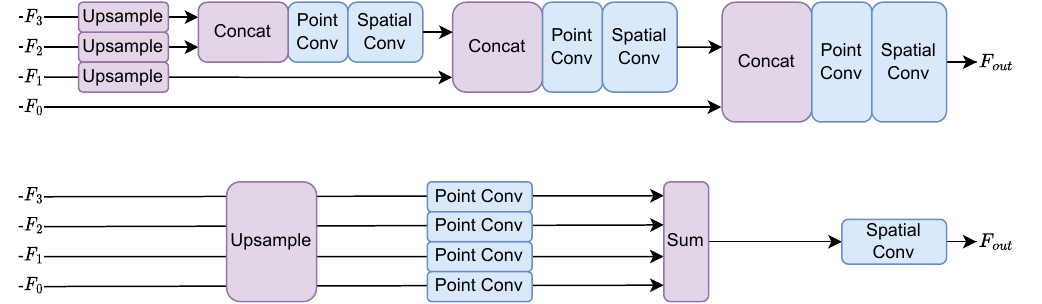}
    \caption{Overview of the two decoders. On top is the pyramid decoder with concatenation as a fusion method. On the bottom is the flat decoder with sum as the fusion method. The input feature maps are ordered from the most shallow to the deeper ones. ``Spatial Conv'' is a $3\times3$ convolution, while ``Point Conv'' is a $1 \times 1$ convolution.}\label{fig:decoders}
\end{figure}

Each decoder is followed by a $1 \times 1$ convolution to reduce the number of channels to the number of classes and a softmax activation function to output the probability of each class. The two decoders are shown in Fig.~\ref{fig:decoders}.

\section{Experimental Evaluation}
\label{sec:experiments}

\subsection{Dataset and Validation}

We assessed RoWeeder on the WeedMap dataset~\cite{sa2018weedmap}, which includes multispectral images from sugar beet fields in Germany (Rheinbach) and Switzerland (Eschikon). These images were captured using UAVs equipped with RedEdge-M and Sequoia cameras, respectively (see~\cite{sa2018weedmap} for further details). The dataset comprises eight orthomosaic maps, labeled [000] through [007], with the first five belonging to the Rheinbach subset and the last three to the Eschikon subset. Multiple tiles of size $512 \times 512$ pixels were derived from each orthomosaic map by sliding a non-overlapping window over the maps.

We focused on the Rheinbach subset because the Eschikon subset lacks the blue channel necessary for our RGB-focused experiments. Experimentation with other channels is beyond the scope of this work. The dataset was split to perform a 5-fold cross-validation. Additionally, the training set was randomly divided into training and validation subsets, and the validation subset was used for model selection. We randomly divided the training set rather than basing it on specific fields to avoid relying solely on a single field for validation. The dataset contains annotations for crops and weeds used only to evaluate the model's performance at test time.

We used the macro-averaged F1 score as a metric for our experiments, which can deal with the class imbalance typical of this task.

\subsection{Setting}

The NDVI-based plant detection system utilized a threshold value of 0.1 to detect significant pixels of plants. For the Hough transform, a threshold of 160 was used. The uniformity test employed a p-value of 0.1. In the SLIC algorithm, the number of clusters was calculated as $n = 0.005 \times H \times W$, where $H$ and $W$ are the image height and width. The compactness parameter was set to 20, and $\sigma$ was set to 1. The mentioned parameters were selected through an empirical evaluation of the detected lines within a representative sample of the dataset.

Our training strategy optimized the focal loss~\cite{lin2017focal}, which addresses class imbalance by focusing on difficult-to-classify examples. The loss for a single training instance was computed as:
\begin{equation}
\mathcal{L} = \frac{1}{N+1}\sum_n^{N+1}{\left[w_n \cdot (1 - e^{-l_{ce}(\hat{y}_n, y_n)})^{\gamma} \cdot l_{ce}(\hat{y}_n, y_n)\right]},
\end{equation}
where $N + 1$ is the number of classes, $w_n$ are class-specific weights, $l_{ce}$ is the cross-entropy loss, and $\gamma$ is the focusing parameter that adjusts the emphasis on hard examples. We used the AdamW optimizer~\cite{loshchilov2017decoupled} with $\beta_1 = 0.9$ and $\beta_2 = 0.999$, complemented by a linear learning rate warmup~\cite{goyal2017accurate} for 1000 iterations, followed by a step-wise cosine learning rate decay schedule~\cite{loshchilov2016sgdr}. The initial learning rate after warmup was $1\mathrm{e}{-5}$. The model was trained for 20 epochs, with the iteration count per epoch equivalent to the size of the training set.

Training harnessed the resources of the Leonardo cluster, which provided 512GB of RAM and four NVIDIA A100-64GB GPUs. This setup ensured the high-performance processing capabilities necessary to support our experiments' computational demands.

\subsection{Results}

We conducted a 5-fold cross-validation on the Rheinbach subset, with each fold corresponding to a different orthomosaic map. The pseudo-ground truth was used for the training folds, and the original ground truth was used for the testing folds. The results in Table \ref{tab:folds} show that our method achieved a mean F1 score of 75.3, outperforming the baseline methods Hough+CC and Hough+SLIC. The results demonstrate that the model can learn to distinguish between crops and weeds even with noise in the pseudo-ground truth. This is particularly evident in fold 004, where the pseudo-ground truth generated by the Hough+CC method yielded an F1 score of 63.0, while our method achieved an F1 score of 74.3, and the Hough+CC+ResNet50 method adapted from~\cite{bahDeepLearningUnsupervised2018} achieved an F1 score of 52.9. Fold 004 contains some low-quality, blurred images that may have compromised the Hough algorithm's predictions and effectiveness. However, the end-to-end model successfully overcame this issue, maintaining its accuracy.
Note that RoWeeder learns from the output of Hough+CC, so achieving a higher score than its ground truth indicates its resilience to noise.
The results also indicate that the SLIC superpixels do not enhance the model's performance, as the Hough+SLIC method achieved the same F1 score as the Hough+CC method.

The noise in the pseudo-ground truth arises from intra-row weeds being misclassified as crops. To evaluate the model's ability to handle this issue, we calculated two F1-scores: one for the pixels of plants within the rows, termed the \textit{intra-row F1-score}, and another for the pixels of plants outside the rows, termed the \textit{inter-row F1-score}. The results, averaged over the five folds test set, are presented in Table \ref{tab:rowtest}.
Given the definitions of these two metrics, the inter-row F1-score for crops is 0, as the model cannot detect crops outside the rows. Similarly, the intra-row F1-score for weeds is 0 since the model fails to identify weeds within the rows. These scores, therefore, reflect the model's ability to learn from noisy data, indicating its noise resiliency.

\begin{table}[t]
  \centering
\begin{tabular}{lcccccc}
\hline
\multicolumn{1}{l|}{\multirow{2}{*}{Method}}               & \multicolumn{5}{c|}{Field}                                                               & \multirow{2}{*}{Mean} \\
\multicolumn{1}{l|}{}                                      & 000           & 001           & 002           & 003           & \multicolumn{1}{c|}{004} &                                 \\ \hline
Hough+SLIC+ResNet50 \cite{bahDeepLearningUnsupervised2018} & 73.8          & \textbf{79.3} & 72.7          & \textbf{77.5} & 52.9                     & 71.2 $\pm$ 4.2                  \\
Hough+CC                                                   & 68.1          & 72.0          & \textbf{76.1}          & 72.7          & 63.0                     & 70.4 $\pm$ 2.0                  \\
Hough+SLIC                                                 & 68.1          & 72.0          & \textbf{76.1} & 72.5          & 63.1                     & 70.4 $\pm$ 2.0                  \\ \hline
RoWeeder (SegFormer)                                       & 70.3          & 76.4          & 75.5          & 77.1          & 72.8                     & 74.4 $\pm$ 1.1                  \\
RoWeeder (Pyramid)                                         & 71.6          & 74.2          & 74.7          & 76.8          & 72.0                     & 73.9 $\pm$ 0.8                  \\
RoWeeder (Flat)                                            & \textbf{74.5} & 74.9          & 75.6          & 77.2          & \textbf{74.3}            & \textbf{75.3 $\pm$ 0.5}         \\ \hline
\end{tabular}
    \caption{F1 scores from 5-fold cross-validation on the Rheinbach subset. RoWeeder, in the three different settings, outperforms the baselines and the competitors.} \label{tab:folds}
\end{table}

\begin{table}[t]
\centering
\begin{tabular}{l|ccc|ccc|ccc}
\hline
\multirow{2}{*}{Method} &
  \multicolumn{3}{c|}{Full} &
  \multicolumn{3}{c|}{Inter-row} &
  \multicolumn{3}{c}{Intra-row} \\
                     & BG   & Crop & Weed & BG   & Crop & Weed & BG   & Crop & Weed \\ \hline
Hough+SLIC+ResNet50 \cite{bahDeepLearningUnsupervised2018} &
  \textbf{98.5} &
  62.6 &
  52.6 &
  \textbf{99.0} &
  25.4 &
  54.6 &
  \textbf{98.8} &
  64.5 &
  35.1 \\
Hough+CC             & \textbf{98.5} & 65.8 & 46.9 & \textbf{99.0} & 01.9 & 55.6 & \textbf{98.8} & 69.4 & 00.1 \\ \hline
RoWeeder (SegFormer) & 98.4 & 69.5 & 55.4 & 98.8 & 40.9 & 59.5 & 98.6 & 70.3 & 38.2 \\
RoWeeder (Pyramid)   & 98.4 & 69.9 & 53.3 & 98.9 & 41.0 & 56.6 & 98.7 & 71.3 & 36.9 \\
RoWeeder (Flat) &
  98.4 &
  \textbf{70.9} &
  \textbf{56.6} &
  98.9 &
  \textbf{42.4} &
  \textbf{59.8} &
  98.7 &
  \textbf{71.9} &
  \textbf{40.3} \\ \hline
\end{tabular}
\caption{Per-class F1 score averaged over 5-fold cross-validation. \textit{Full} represents the F1 score calculated across all pixels in the image. \textit{Inter-row} refers to the F1 score for pixels belonging to plants on the crop rows, while \textit{inter-rows} refers to the F1 score for pixels belonging to plants outside the crop rows. \textit{BG} stands for background.}\label{tab:rowtest}
\end{table}

Table \ref{tab:results} presents the computational requirements of the analyzed methods. The Hough+SLIC+ResNet50 method uses a CNN to classify detected crop rows and requires significant computational resources, with an inference time of 1620ms. In contrast, our method achieves an inference time of 7ms, making it suitable for real-time applications through drones. Inference time was calculated on a single NVIDIA RTX 4090, a consumer-grade GPU, which is more powerful than a typical edge device but can still reflect the difference in computational cost between the methods. The RoWeeder model is lightweight, with only 3.6M parameters and 3.68 GMACs, making it suitable for edge-device deployment. Future work will focus on testing the model on edge devices to evaluate its performance in real-world scenarios. 

\begin{table}[t]
  \centering
  \begin{tabular}{lccc}
    \hline
    Method                                                     & Params (M)    & GMACs         & Inference time (ms) \\ \hline
    Hough+SLIC+ResNet50 \cite{bahDeepLearningUnsupervised2018} & 23.51         & 21.58*        & 1620                \\
    Hough+CC                                                   & /             & /             & 84                  \\
    Hough+SLIC                                                 & /             & /             & 377                 \\ \hline
    RoWeeder (SegFormer)                                       & 3.71          & 7.84          & \textbf{7.07}       \\
    RoWeeder (Pyramid)                                         & 3.90          & 3.91          & 7.79                \\
    RoWeeder (Flat)                                            & \textbf{3.60} & \textbf{3.68} & 7.16                \\ \hline
  \end{tabular}
  \caption{Computational cost comparison between RoWeeder and competitors. * indicates that GMACs are calculated for the deep learning model only, and the method needs to rerun the deep learning model for each plant in the image, which is variable.}
  \label{tab:results}
\end{table}

\begin{figure}[t]
    \centering
    \begin{subfigure}{0.23\linewidth}
      \includegraphics[width=\linewidth]{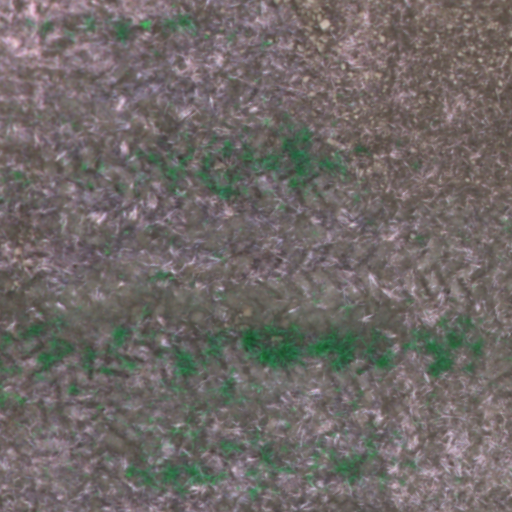}
    \end{subfigure}
    \begin{subfigure}{0.23\linewidth}
      \includegraphics[width=\linewidth]{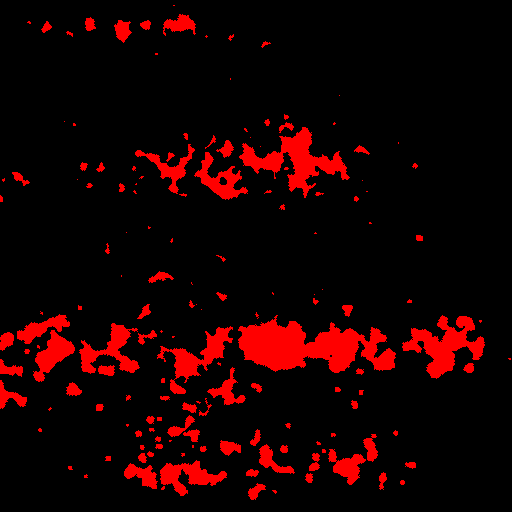}
    \end{subfigure}
    \begin{subfigure}{0.23\linewidth}
        \includegraphics[width=\linewidth]{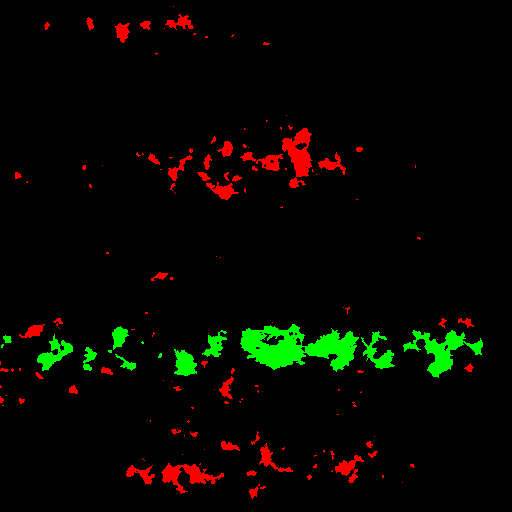}
    \end{subfigure}
    \begin{subfigure}{0.23\linewidth}
      \includegraphics[width=\linewidth]{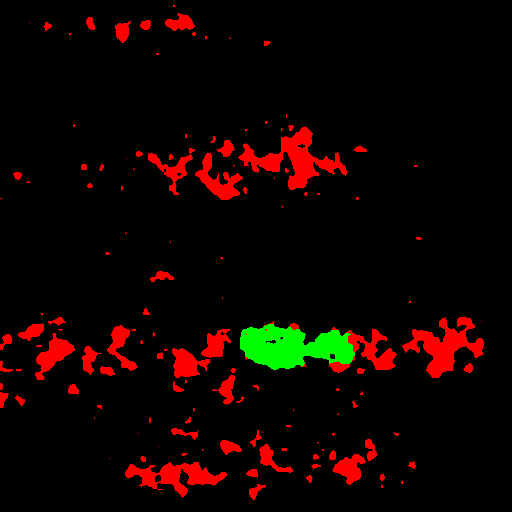}
    \end{subfigure}
    
    \begin{subfigure}{0.23\linewidth}
      \includegraphics[width=\linewidth]{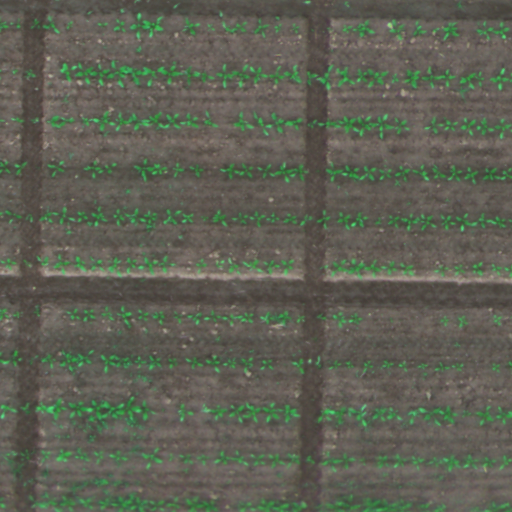}
    \end{subfigure}
    \begin{subfigure}{0.23\linewidth}
      \includegraphics[width=\linewidth]{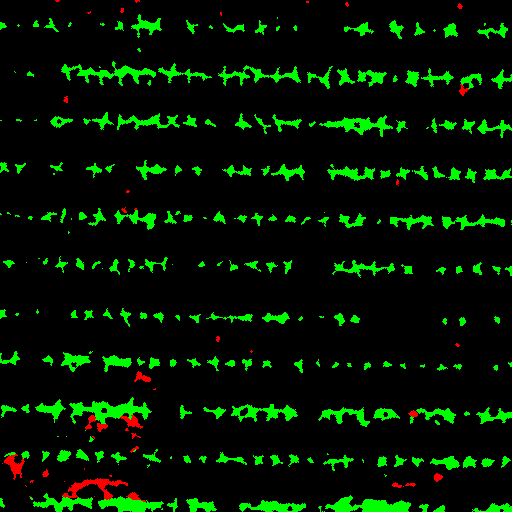}
    \end{subfigure}
    \begin{subfigure}{0.23\linewidth}
        \includegraphics[width=\linewidth]{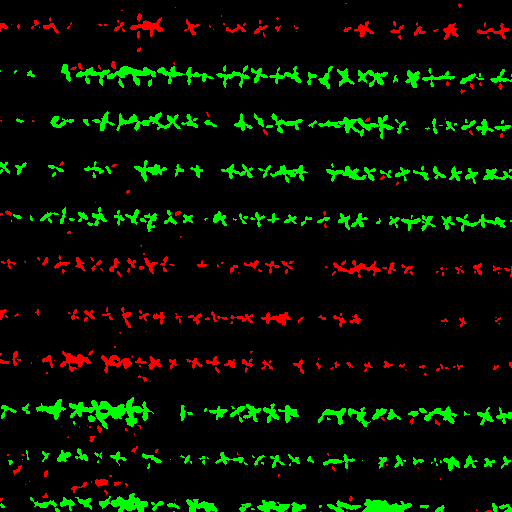}
    \end{subfigure}
    \begin{subfigure}{0.23\linewidth}
      \includegraphics[width=\linewidth]{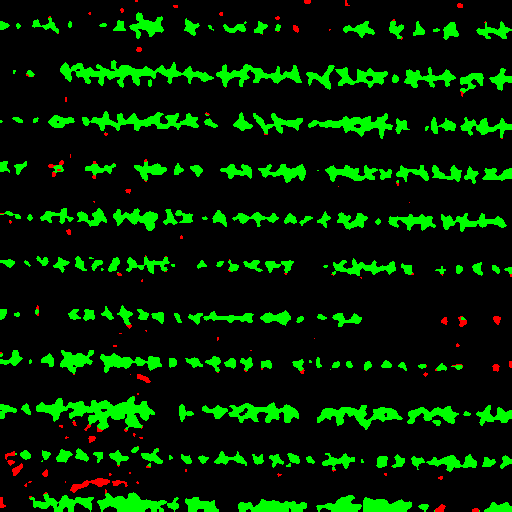}      
    \end{subfigure}

    \begin{subfigure}{0.23\linewidth}
      \includegraphics[width=\linewidth]{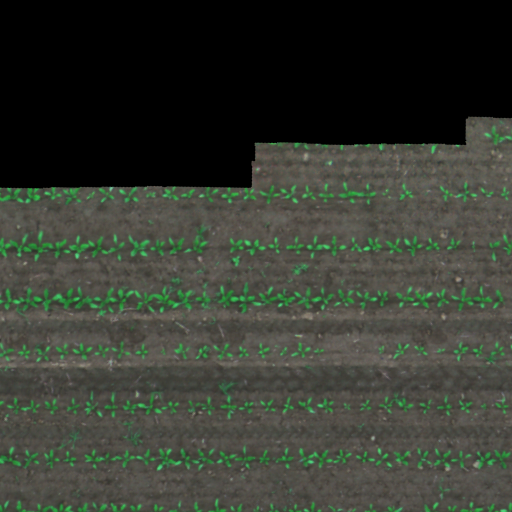}
    \end{subfigure}
    \begin{subfigure}{0.23\linewidth}
      \includegraphics[width=\linewidth]{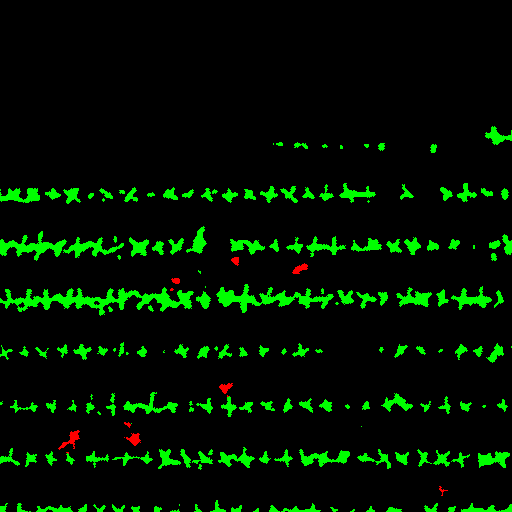}
    \end{subfigure}
    \begin{subfigure}{0.23\linewidth}
        \includegraphics[width=\linewidth]{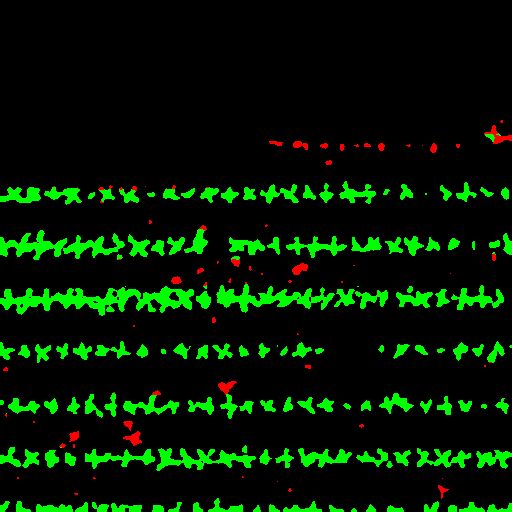}
    \end{subfigure}
    \begin{subfigure}{0.23\linewidth}
      \includegraphics[width=\linewidth]{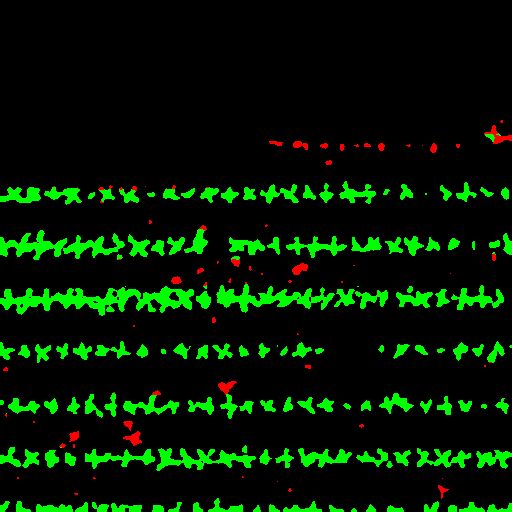}      
    \end{subfigure}

    \begin{subfigure}{0.23\linewidth}
      \includegraphics[width=\linewidth]{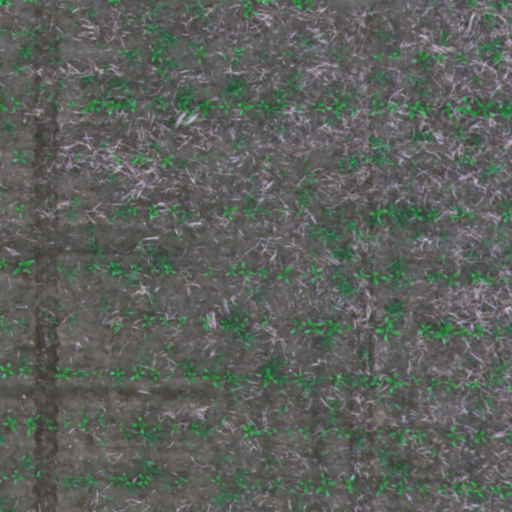}
      \captionsetup{labelformat=empty}
      \caption{Input image}
    \end{subfigure}
    \begin{subfigure}{0.23\linewidth}
      \includegraphics[width=\linewidth]{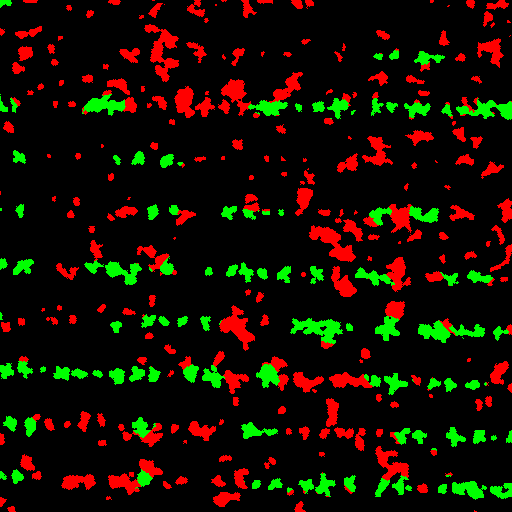}
      \captionsetup{labelformat=empty}
      \caption{Ground truth}
    \end{subfigure}
    \begin{subfigure}{0.23\linewidth}
        \includegraphics[width=\linewidth]{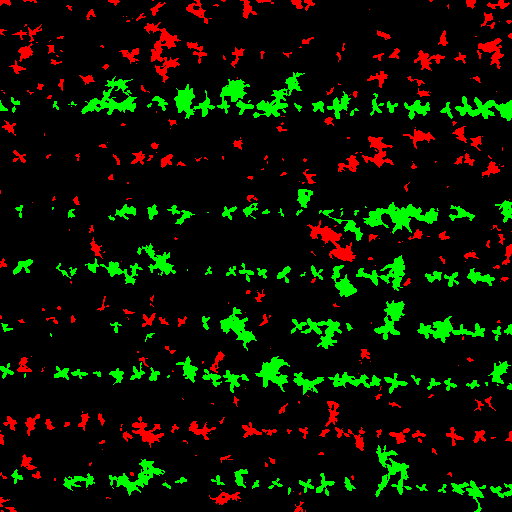}
        \captionsetup{labelformat=empty}
        \caption{Pseudo-ground truth}
    \end{subfigure}
    \begin{subfigure}{0.23\linewidth}
    \captionsetup{labelformat=empty}
      \includegraphics[width=\linewidth]{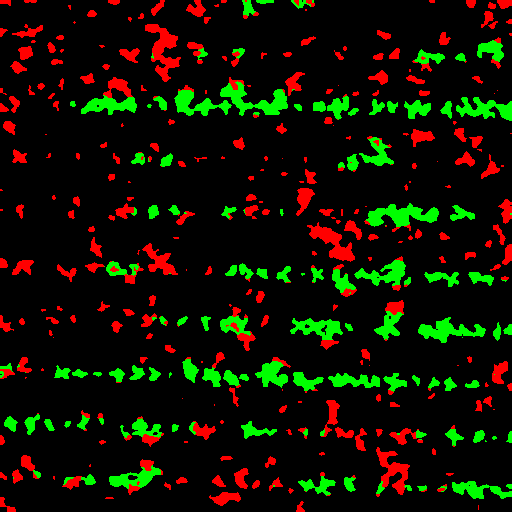}      
      \caption{Prediction}
    \end{subfigure}

    \caption{Pseudo-ground truth is generated with the Hough+CC method and prediction by RoWeeder. The background is denoted in black, crops in green, and weeds in red.}
    \label{fig:qresults}
\end{figure}

Figure \ref{fig:qresults} presents qualitative results from the RoWeeder (flat) model. The first two rows highlight the primary limitation of the Hough baselines. In the first example, a line of weeds is misidentified as a crop row, while in the second, some crop rows go undetected. These errors stem from the Hough transform's sensitivity to parameters, such as the threshold value. Applying uniform parameters across an entire dataset can lead to such inaccuracies. Despite these ground truth errors, RoWeeder demonstrates robust performance, successfully identifying most weeds in the first example and most crops in the second.

\subsection{Ablation Study}

Our comprehensive ablation studies examined different components and configurations of the RoWeeder decoders. These experiments were performed using field 003 as the test set. Table \ref{tab:ablations} shows the results for different fusion and upsampling methods across the two decoder settings. The results indicate that the flat decoder with sum as the fusion method achieved the best performance, with an F1 score of 77.2. The pyramid decoder with concatenation as the fusion method was the second best, with an F1 score of 76.8. Thus, for each decoder, a different fusion method proved more suitable. Additionally, upsampling through interpolation yielded better results than deconvolution.

Another component evaluated was the decoder's $3 \times 3$ spatial convolution. An ablation study over field 003, removing the spatial convolution from the flat and pyramid decoders, showed that spatial convolution is crucial for good performance. The F1 score dropped from 77.2 to 75.4 for the flat decoder and from 76.8 to 74.0 for the pyramid decoder when the spatial convolution was removed. Lastly, we evaluated the impact of the number of blocks in the model. Table \ref{tab:blocks} shows that removing blocks from the encoder and decoder decreased the model's performance for both the flat and pyramid decoders. However, GMACs linearly decrease with the number of blocks, while the F1 score drop is less pronounced. Therefore, this trade-off could be considered when the environment has limited computational resources.

We also trained our model in a supervised setting to establish a benchmark for the unsupervised approach. Using 5-fold cross-validation, we achieved an F1 score of 82.8. Although the unsupervised method results in a 7.5-point decrease in F1 score, it still delivers competitive performance.

\begin{table}[t]
  \centering
    \begin{tabular}{ll|cc}
    \hline
    \multirow{2}{*}{Upsample}      & \multirow{2}{*}{Fusion} & \multicolumn{2}{c}{F1 score}                           \\
                                   &                         & \multicolumn{1}{l}{Pyramid} & \multicolumn{1}{l}{Flat} \\ \hline
    \multirow{2}{*}{Interpolation} & Add                     & 73.7                        & \textbf{77.2}            \\
                                   & Concat                  & \textbf{76.8}               & 75.3                     \\ \hline
    \multirow{2}{*}{Deconvolution} & Add                     & 73.1                        & 73.1                     \\
                                   & Concat                  & 74.7                        & 73.5                     \\ \hline
    \end{tabular}
  \caption{Results of different settings of the decoders over field 003.}
  \label{tab:ablations}
\end{table}

\begin{table}[t]
\centering
\begin{tabular}{lc|ccc}
\hline
Method                   & \# blocks & F1 Score      & GMACs         & Params        \\ \hline
\multirow{3}{*}{Pyramid} & 2         & 75.3          & \textbf{2.00} & \textbf{0.69} \\
                         & 3         & 76.3          & 3.10          & 1.67          \\
                         & 4         & \textbf{76.8} & 3.91          & 3.90          \\ \hline
\multirow{3}{*}{Flat}    & 2         & 72.3          & \textbf{2.00} & \textbf{0.69} \\
                         & 3         & 74.8          & 2.97          & 1.62          \\
                         & 4         & \textbf{77.2} & 3.68          & 3.60          \\ \hline
\end{tabular}
\caption{Ablation study over the number of blocks in the encoder and decoder.}
\label{tab:blocks}
\end{table}

\section{Conclusion}
\label{sec:conclusion}

In this study, we presented RoWeeder, an innovative framework for unsupervised weed mapping that combines crop-row detection with a noise-resilient deep learning model. Our approach leverages crop-row information to create a pseudo-ground truth, which is then used to train a deep learning model. We evaluated our method on the WeedMap dataset, achieving an F1 score of 75.3, outperforming other methods. Our results demonstrate that the model can effectively distinguish between crops and weeds, even with noise in the pseudo-ground truth. Additionally, we conducted an ablation study on different components and configurations of the RoWeeder decoders, showing that the flat decoder with sum as the fusion method achieved the best performance.

We envision a broad spectrum of future research directions to advance our method. These include exploring different architectures for the deep learning model, investigating the impact of various crop-row detection methods, and extending the framework to other crops and datasets. We also plan to develop a new decoder more resilient to noise by leveraging contrastive learning to build a class prototype for each class and then use this prototype to classify each pixel.

\subsubsection{Acknowledgement} 

We acknowledge the CINECA award under the ISCRA initiative for providing us with access to high-performance computing resources and support. The research of Pasquale De Marinis is funded by a Ph.D.~fellowship within the framework of the Italian ``D.M.~n.~352, April 9, 2022'' - under the National Recovery and Resilience Plan, Mission 4, Component 2, Investment 3.3 - Ph.D.~Project ``Computer Vision techniques for sustainable AI applications using drones'', co-supported by ``Exprivia S.p.A.'' (CUP H91I22000410007).